\documentclass{article}
\usepackage[table]{xcolor}
\usepackage[OT1]{fontenc}
\usepackage{iclr2027_conference,times}
\usepackage{microtype}
\usepackage{graphicx}
\graphicspath{{fig/}{../fig/}{main/fig/}}
\usepackage{subcaption}
\usepackage{booktabs}
\usepackage{threeparttable}
\usepackage{longtable}
\usepackage{multirow}
\usepackage{float}
\usepackage{algorithm}
\usepackage{algpseudocode}
\usepackage{wrapfig}
\usepackage{needspace}
\makeatletter
\g@addto@macro\fancy@reset{%
  \everypar{}%
  \let\parshape\@@parshape\parshape=0\relax
  \let\par\@@par\let\@par\@@par\let\@setpar\@@setpar}
\makeatother
\usepackage{array}
\usepackage{listings}
\usepackage{amsmath,amssymb,mathtools,amsthm}
\usepackage{hyperref}
\definecolor{citationblue}{RGB}{37,102,151}
\definecolor{referencebrick}{RGB}{148,57,49}
\hypersetup{colorlinks=true,citecolor=citationblue,linkcolor=referencebrick,urlcolor=citationblue}
\hypersetup{
  pdftitle={CLIVO: Differentiable Cluster-Based Instruction Mixing with Learned Validation Objectives},
  pdfauthor={Yuhan Wang, Rui Hu, Batian Di, Tan Jiang, Dongsheng Ma, Leichao Dong, Yuanzi Li, Yalin Feng},
  pdfsubject={Target-aware multimodal instruction data selection}
}
\usepackage{url}
\usepackage[capitalize,noabbrev]{cleveref}
\crefname{section}{Section}{Sections}
\Crefname{section}{Section}{Sections}
\crefname{subsection}{Section}{Sections}
\Crefname{subsection}{Section}{Sections}
\crefname{subsubsection}{Section}{Sections}
\Crefname{subsubsection}{Section}{Sections}

\theoremstyle{plain}

\theoremstyle{definition}

\theoremstyle{remark}

\title{CLIVO: Differentiable Cluster-Based\\Instruction Mixing with Learned\\Validation Objectives}
\author{\parbox[t]{\dimexpr\textwidth-2\tabcolsep\relax}{\centering\bfseries
Yuhan Wang\textsuperscript{1}\quad Rui Hu\textsuperscript{1}\quad Batian Di\textsuperscript{2}\quad Tan Jiang\textsuperscript{3}\\
Dongsheng Ma\textsuperscript{1}\quad Leichao Dong\textsuperscript{4}\quad Yuanzi Li\textsuperscript{2}\quad Yalin Feng\textsuperscript{5}\\[4pt]
\normalfont\textsuperscript{1}Peking University \quad
\textsuperscript{2}Renmin University of China \quad
\textsuperscript{3}Zhejiang University\\
\normalfont\textsuperscript{4}Xi'an Jiaotong University \quad
\textsuperscript{5}Nanyang Technological University}}
\iclrfinalcopy

\begin{document}
\flushbottom
\maketitle
\fancyhead{}
\renewcommand{\headrulewidth}{0pt}

\begin{abstract}
Data selection turns large multimodal instruction pools into compact training subsets tailored to a vision-language model's target capabilities. Effective selection requires both useful examples and an appropriate balance of supervision. Sample-level rankings can overrepresent similar content without optimizing this balance, while group-level methods are constrained by coarse allocation units, predefined validation sets, and costly or inaccurate mixture search. We propose CLIVO, a target-aware framework that connects fine-grained semantic allocation, automatically learned validation objectives, and differentiable mixture search. Specifically, semantic clusters define independently adjustable training shares, and merged cluster experts provide reusable proxies for evaluating mixtures. We select validation examples by correlating their proxy losses with real probe performance, then learn nonnegative weights to form a target-specific objective. With experts and validation weights fixed, a single backward pass supplies gradients for all shares, avoiding an additional mixture-response regression model. Selecting 100K instructions from an 8M pool, CLIVO attains 97.2\% of the full-data four-target average score using only 1.25\% of the data. At this budget, it outperforms the strongest evaluated selection baseline by 1.16 points with random within-cluster sampling; combining learned quotas with sample ranking increases this advantage to 1.55 points. CLIVO further supports interpretable, flexible, and reusable data curation through semantic visualization, modular selectors, and shared experts.
\end{abstract}

\section{Introduction}
\label{sec:introduction}

Vision-language models (VLMs) have demonstrated strong capabilities in visual understanding, document parsing, multimodal reasoning, and interactive tasks~\citep{comanici2025gemini,wang2025internvl3,bai2025qwen3,hong2025glm,huang2026step3,team2026kimi}.
Their development combines multimodal representation pretraining with instruction-following post-training~\citep{yin2025sail,team2025longcat,team2025kimi,radhakrishna2025apriel,lin2024vila}.
Multimodal instruction tuning develops instruction following through instruction--response examples about visual inputs~\citep{liu2023visual,dai2023instructblip,du2025makes}.
The training data therefore shape both the supervision a model receives and the tasks it practices. Effective tuning depends on supervision quality~\citep{luo2025mmevol} as well as the composition of the training set~\citep{zhang2025mm1}.
Large instruction pools offer broad coverage of tasks and response formats~\citep{xu2024vision}, yet this breadth does not ensure that every example is useful: incorrect responses and hallucinated content can degrade model performance~\citep{gou2025corrupted}.
This motivates \textbf{data selection}, which concentrates training on useful supervision. Prior work shows that carefully selected subsets can match full-data performance and even surpass it on individual benchmarks~\citep{liu2025less}.
Compact subsets also support repeated adaptation to different applications. A document assistant, for example, needs a different balance of text reading, visual recognition, and reasoning from a general visual assistant.
Among useful examples, relative abundance determines how much training each type of supervision receives. Selection must therefore address both which examples to retain and how to balance their contribution to the intended capability.
At a fixed training-data budget, we ask:

\noindent\textbf{How should instruction data be mixed to best support a target capability?}

Existing data selection methods operate at the sample or group level.
\emph{Sample-level methods} rank examples using signals such as gradient influence in ICONS~\citep{wu2024icons}, necessity in MLLM-Selector~\citep{ma2025mllm}, and importance and diversity in CoIDO~\citep{yan2026coido}.
However, selecting the highest-scoring examples can overrepresent similar content, reducing diversity and degrading downstream performance~\citep{he2025learning}.
Diversity-aware scores help, but often leave content shares implicit.
\emph{Group-level methods} explicitly allocate shares across sources, domains, or semantic groups, requiring choices of allocation \textbf{unit}, evaluation \textbf{signal}, and mixture \textbf{learning} rule.

\noindent\textbf{Unit: fine-grained allocation remains difficult to search.}
Predefined source shares tie heterogeneous instructions together.
Semantic clusters offer finer control, but COINCIDE's density and transferability heuristics~\citep{lee2024concept} do not optimize measured target performance.
CLIMB~\citep{diao2025climb} uses downstream feedback to search 21 groups in its main setting, with sampling costs limiting substantially finer search.

\noindent\textbf{Signal: validation relies on predefined sets and coarse loss averages.}
Costly candidate training motivates proxy validation losses.
CLIMB uses predefined task sets, while CAMEL~\citep{li2026capacity} maps validation losses to benchmark scores.
Neither automatically selects target-relevant examples, and their set-level averages can obscure informative differences between individual responses.

\noindent\textbf{Learning: search is costly and susceptible to response-prediction error.}
DoReMi~\citep{xie2023doremi} couples reweighting to proxy training, limiting reuse across targets and groupings.
RegMix~\citep{liu2024regmix}, DeMix~\citep{li2026decouple}, and MergeMix~\citep{wang2026mergemix} instead fit response predictors from evaluated mixtures.
High-dimensional candidate labeling is costly, and selecting predicted optima can amplify errors on unseen mixtures~\citep{smith2006optimizer}.

We propose \textbf{CLIVO} (\textbf{CL}uster-based \textbf{I}nstruction Mixing with Learned \textbf{V}alidation \textbf{O}bjectives) to jointly address these challenges. Specifically, \textbf{fine-grained semantic units} provide adjustable shares at the hundred-cluster scale. A \textbf{learned validation signal} selects examples using proxy-loss correlations with real probe scores and fits nonnegative weights to aggregate their losses. Expert merging supplies proxies without retraining for each mixture~\citep{berasi2026linear}. Finally, \textbf{gradient-based mixture learning} differentiates this objective through fixed experts to update all shares jointly, without an additional mixture-response predictor.

Using 100K instructions (1.25\% of the pool) and random within-cluster sampling, CLIVO attains 97.2\% of the full-data four-target average score.
It improves on the strongest evaluated data selection baseline at this budget by 1.16 points; within-quota ranking increases the advantage to 1.55 points.
Our analyses show that finer units benefit most targets, moderate-budget experts and sample weighting provide reliable feedback while preserving individual responses, and direct gradient search lowers every target's learned objective.

Importantly, CLIVO also supports \textbf{interpretable, flexible, and reusable data curation}.
Semantic visualizations expose content choices; modular encoders, objectives, and selectors accommodate different applications.
With the base and partition fixed, experts, probes, and cached losses support new targets.
A single backward pass supplies gradients for all shares even at finer granularities, avoiding an additional higher-dimensional regression fit.

Our contributions are:
\begin{itemize}
\setlength{\itemsep}{2pt}
\setlength{\parsep}{0pt}
\item \textbf{A unified target-aware framework.} We connect fine-grained semantic allocation, automatic validation-example selection and weighting, and differentiable mixture search through reusable expert proxies.
\item \textbf{Evidence for fine-grained allocation, reliable signals, and direct search.} We demonstrate benefits from finer units on most targets, reliable feedback from moderate-budget experts and individual validation responses, and objective improvement through direct search for every target.
\item \textbf{Interpretable, flexible, and reusable curation.} We make allocations interpretable through visualization, demonstrate flexible combinations with sample selectors, and reuse experts and feedback across targets and granularities.
\end{itemize}

\begin{figure}[!t]
\centering
\includegraphics[width=\linewidth]{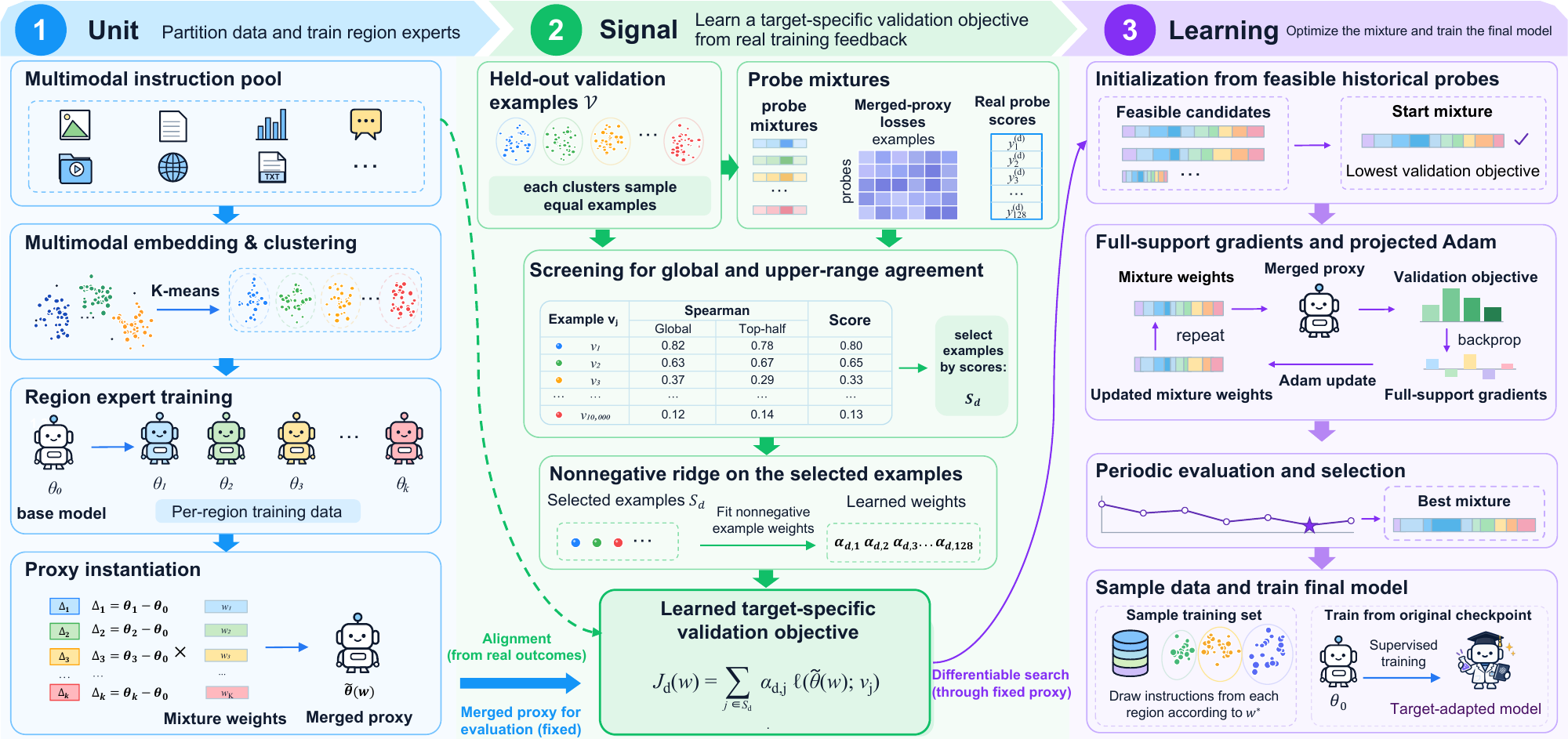}
\caption{\textbf{CLIVO overview: learning what to train on and how to evaluate the mixture.} \textbf{(1) Unit:} a fixed multimodal encoder and $K$-means organize the instruction pool into semantic clusters with adjustable training shares. \textbf{(2) Signal:} probe models are trained on sampled mixtures and evaluated on downstream benchmarks. Cluster experts are merged using the same shares to construct proxies. Correlations between per-example proxy losses and real scores identify target-relevant validation examples; learned nonnegative weights combine their losses into an objective. \textbf{(3) Learning:} with experts and validation weights frozen, gradients update the training shares through the merged proxy under budget and capacity constraints. Periodic evaluations select the lowest-loss mixture. Its quotas specify how many examples to sample from each cluster for training from the original base.}
\label{fig:pipeline}
\end{figure}

\section{Method}
\label{sec:method}

\paragraph{Framework overview.}
CLIVO learns target-specific training shares through three decisions: \emph{how to group the data}, \emph{how to evaluate a mixture}, and \emph{how to update its shares} (\Cref{fig:pipeline,alg:clivo}). We define the task (\Cref{sec:problem-setup}), then present semantic units (\Cref{sec:semantic-allocation}), validation learning (\Cref{sec:allocation-signal}), and gradient search (\Cref{sec:allocation-search}).

\subsection{Problem Setup}
\label{sec:problem-setup}

Given an instruction pool $\mathcal D$, an initial VLM $\theta_0$, a target capability $d$, and a budget $N$, we seek a size-$N$ subset maximizing the target score after instruction tuning. For a partition $\{\mathcal D_k\}_{k=1}^K$, we optimize admissible training shares $\mathbf w=(w_1,\ldots,w_K)\in\mathcal W$, with $w_k\geq0$ and $\sum_kw_k=1$. These shares determine sample counts $n_k(\mathbf w)\approx Nw_k$ totaling $N$. We fix the within-group rule $\pi_{\rm U}$ to uniform sampling without replacement, yielding $\mathcal S(\mathbf w,\pi_{\rm U})=\bigcup_k\pi_{\rm U}(\mathcal D_k,n_k(\mathbf w))$. Let $\theta_{\mathcal S}=\operatorname{Train}(\theta_0,\mathcal S)$ denote the model trained with a fixed recipe and $R_d(\theta)$ its target score. Our objective is
\begin{equation}
\mathbf w_d^\star\in\arg\max_{\mathbf w\in\mathcal W}F_d(\mathbf w;\pi_{\rm U}),
\qquad
F_d(\mathbf w;\pi_{\rm U})=\mathbb E_{\mathcal S}
\left[R_d(\theta_{\mathcal S})\right].
\label{eq:selection-task}
\end{equation}
The expectation accounts for score variation across random subsets drawn with the same shares.

\Needspace*{45pt}
\subsection{Unit: Semantic Allocation}
\label{sec:semantic-allocation}

\setlength{\intextsep}{0pt}
\leavevmode\begin{wrapfloat}{algorithm}{R}{0.58\textwidth}
\centering\small
\captionsetup{font=small,skip=4pt}
\caption{CLIVO: target-specific instruction mixing}
\label{alg:clivo}
\begin{minipage}{\linewidth}
\begin{algorithmic}
\Require $\mathcal D,\theta_0,E,R_d,N$; $K,m,T,\{\mathbf w_i\}_{i=1}^{Y}$
\Ensure Shares $\widehat{\mathbf w}_d$; subset $\mathcal S_d^\star$
\Statex \textbf{1. Unit: semantic allocation}
\Statex \hspace{\algorithmicindent}Encode $\mathcal D$ with fixed external encoder $E$.
\Statex \hspace{\algorithmicindent}Apply $K$-means; obtain clusters $\{\mathcal D_k\}$.
\Statex \textbf{2. Signal: learned validation objective}
\Statex \hspace{\algorithmicindent}Train probes at $\{\mathbf w_i\}$; evaluate $\mathbf y^{(d)}$.
\Statex \hspace{\algorithmicindent}Train cluster experts $\{\theta_k\}$ from $\theta_0$.
\Statex \hspace{\algorithmicindent}$\mathcal V\gets\operatorname{BalancedSample}(\{\mathcal D_k\})$
\Statex \hspace{\algorithmicindent}$X_{ij}\gets-\ell(\widetilde\theta(\mathbf w_i);v_j)$
\Statex \hspace{\algorithmicindent}$S_d\gets\operatorname{Top}_m((\rho_{\rm all,j}+\rho_{\rm upper\ half,j})/2)$
\Statex \hspace{\algorithmicindent}Fit Eq.~\eqref{eq:nonnegative-ridge}; rescale and normalize to $\boldsymbol\alpha_d$.
\Statex \textbf{3. Learning: mixture search}
\Statex \hspace{\algorithmicindent}Freeze $\{\theta_k\},\boldsymbol\alpha_d$.
\Statex \hspace{\algorithmicindent}$\mathbf w^0\gets\arg\min_{\mathbf w_i\in\mathcal W}J_d(\mathbf w_i)$; $\widehat{\mathbf w}_d\gets\mathbf w^0$
\Statex \hspace{\algorithmicindent}\textbf{for} $t=1,\ldots,T$ \textbf{do}
\Statex \hspace{\algorithmicindent}\hspace{\algorithmicindent}$\mathbf g^t\gets\nabla_{\mathbf w}J_d(\mathbf w^{t-1})$ by Eq.~\eqref{eq:mixture-gradient}
\Statex \hspace{\algorithmicindent}\hspace{\algorithmicindent}$\mathbf w^t\gets\Pi_{\mathcal W}(\operatorname{Adam}(\mathbf w^{t-1},\mathbf g^t))$
\Statex \hspace{\algorithmicindent}\hspace{\algorithmicindent}Update best $\widehat{\mathbf w}_d$ at checkpoints.
\Statex \hspace{\algorithmicindent}\textbf{end for}
\Statex \hspace{\algorithmicindent}Convert $\widehat{\mathbf w}_d$ to sample counts $\mathbf n$ totaling $N$.
\Statex \hspace{\algorithmicindent}$\mathcal S_d^\star\gets\bigcup_k\pi_{\rm U}(\mathcal D_k,n_k)$
\end{algorithmic}
\end{minipage}
\end{wrapfloat}
\noindent
Predefined groups rely on source labels or hand-crafted task rules. Semantic clustering instead provides unsupervised allocation units whose granularity is controlled by the number of clusters. We therefore encode instructions with a fixed external multimodal encoder $E$, independent of the VLM being tuned, and apply $K$-means~\citep{mcqueen1967some} to obtain $\{\mathcal D_k\}_{k=1}^K$. Distinct content within a source can receive different shares, while related content across sources can be allocated together. For nested partitions, independent fine-cluster shares retain every coarse allocation and admit additional compositions, expanding the optimization space. The best attainable target score is therefore no lower and can improve when the additional freedom admits a better mixture (\Cref{app:unit-theory}). This flexibility also creates a higher-dimensional search problem, making candidate-by-candidate evaluation costly. We address this with a learned validation objective on reusable proxies (\Cref{sec:allocation-signal}) and gradients that update all shares jointly (\Cref{sec:allocation-search}).

\subsection{Signal: Learning the Validation Objective}
\label{sec:allocation-signal}

\paragraph{Real-training feedback for validation selection.}
We train $Y$ probe models from $\theta_0$ using mixtures $\mathbf w_i$. Uniform within-cluster sampling fixes the selection rule so that the probes test changes in composition (\Cref{app:random-probes}). Their scores $y_i^{(d)}$ supervise validation selection and weighting.

\paragraph{Proxy losses from merged experts.}
Training a model for every candidate mixture is costly. Following prior work~\citep{li2026decouple,wang2026mergemix}, we approximate its losses by merging cluster experts $\theta_k$ trained from $\theta_0$. The candidate's shares determine how these fixed experts are combined. With $\Delta_k=\theta_k-\theta_0$, constructing a proxy requires no further training:
\par
\WFclear
\vspace{2\baselineskip}
\begin{equation}
\widetilde\theta(\mathbf w)
=\theta_0+\sum_{k=1}^{K}w_k\Delta_k
=\sum_{k=1}^{K}w_k\theta_k .
\label{eq:merged-proxy}
\end{equation}

We form a diverse validation candidate pool $\mathcal V=\{v_j\}_{j=1}^{H}$ by uniformly sampling equal numbers of examples from every cluster. For the $Y$ real-probe mixtures, we construct proxies and compute each candidate example's negative loss, $X_{ij}=-\ell(\widetilde\theta(\mathbf w_i);v_j)$, where $\ell$ is the mean cross-entropy over its answer tokens. We use these losses and real scores to identify target-relevant validation examples.

\paragraph{Downstream-aligned weighted signals.}
Search favors low-loss mixtures, so the signal should distinguish strong candidates while ranking the full candidate set. For each example, we average two Spearman correlations between negative loss and target score: one over all fitting probes and one over the half with the highest real scores. Since examples greatly outnumber probes, we retain the top $m$ examples as $S_d$ to limit the number of fitted coefficients~\citep{fan2008sure}. To rank mixtures, we fit centered target scores with nonnegative ridge regression~\citep{hoerl1970ridge}:
\begin{equation}
\widehat{\boldsymbol\beta}_d
=\arg\min_{\boldsymbol\beta\geq0}
\left\|\widetilde X_{\mathcal I,S_d}\boldsymbol\beta
-(\mathbf y_{\mathcal I}^{(d)}-\bar y_d\mathbf1)\right\|_2^2
+\lambda\|\boldsymbol\beta\|_2^2 .
\label{eq:nonnegative-ridge}
\end{equation}
Here $\mathcal I$ indexes all fitting probes, $\widetilde X$ contains their standardized negative losses, $\bar y_d$ is their mean target score, and $\lambda$ regularizes the coefficients. Nonnegative coefficients make lower example losses contribute to higher predicted scores. Rescaling the coefficients to raw-loss units and normalizing them gives weights $\alpha_{d,j}\geq0$ summing to one and the objective
\begin{equation}
J_d(\mathbf w)=\mathcal L_d(\widetilde\theta(\mathbf w))
=\sum_{j\in S_d}\alpha_{d,j}\,
\ell(\widetilde\theta(\mathbf w);v_j).
\label{eq:search-objective}
\end{equation}
Individual weights preserve differences in how examples' losses respond to a mixture, which cluster averaging would remove before fitting. Lower $J_d$ corresponds to a higher predicted score, so the conversion preserves the fitted ranking. The validation weights $\boldsymbol\alpha_d$ determine how mixtures are evaluated; the training shares $\mathbf w$ determine their composition. Detailed settings and the weight conversion are given in \Cref{app:objective-construction}.

\setlength{\intextsep}{8pt plus 2pt minus 2pt}
\subsection{Learning: Gradient-Based Mixture Search}
\label{sec:allocation-search}

The differentiable objective above enables direct, efficient gradient search over the mixture shares. We freeze the experts $\theta_k$ and validation weights $\boldsymbol\alpha_d$, and initialize $\mathbf w^0$ with the lowest-loss previously evaluated feasible mixture to accelerate convergence. At each step, all positive-weight validation examples contribute to the merged proxy's loss. Applying the chain rule~\citep{baydin2018automatic} gives
\begin{equation}
\frac{\partial J_d}{\partial w_k}
=\left\langle
\left.\nabla_\theta\mathcal L_d(\theta)\right|_{\theta=\widetilde\theta(\mathbf w)},
\Delta_k\right\rangle .
\label{eq:mixture-gradient}
\end{equation}
Taking inner products of the accumulated parameter gradient with all $K$
expert differences gives the gradients of every share. Adam updates the
shares jointly while the experts remain fixed, so no separate backward
pass is needed for each coordinate. Periodic evaluations of the full
objective select the lowest-loss iterate $\widehat{\mathbf w}_d$ among
the evaluated checkpoints, including the start.

Finally, we convert the shares into integer counts totaling $N$, sample uniformly within clusters, and train from $\theta_0$ on $\mathcal S(\widehat{\mathbf w}_d,\pi_{\rm U})$. \Cref{app:training-evaluation} gives search and sampling details.

\section{Experiments}
\label{sec:experiments}

In this section, we describe the experimental setup (\Cref{sec:experimental-setup}), compare fixed-budget selection quality (\Cref{sec:main-results}), and conduct ablation studies of CLIVO's components (\Cref{sec:component-results}).

\subsection{Experimental Setup}
\label{sec:experimental-setup}

\paragraph{Model and evaluation.}
We build a 7B VLM with a Qwen2.5-VL-style architecture~\citep{Bai2025Qwen25VLTR}, following the multimodal training approach of LLaVA-1.5~\citep{liu2024improved}. Alignment and mid-training produce the common base $\theta_0$ for all experts, probes, and final models; construction details are in \Cref{app:model-construction}. Evaluation covers General visual understanding (G: MMBench-EN, MMStar, GQA), Parsing (P: OCRBench, ChartQA), and Knowledge and text-centric reasoning (K: MMLU, MMLU-Pro). Each domain's benchmark mean defines a target; Overall (O) averages all seven benchmarks. We learn one mixture per target, and Average summarizes their four scores. Evaluation protocols are in \Cref{app:training-evaluation}.

\paragraph{Data pool.}
We collect approximately 40M instructions from public datasets and their derivatives, covering diverse visual content, tasks, and response formats. To keep selection costs manageable, experiments use an 8M candidate pool. All selection methods draw 100K instructions from this pool; Full uses all 8M. \Cref{app:data-sources} provides data statistics.

\paragraph{Baselines.}
For \emph{sample-level selection}, we choose PPL~\citep{jelinek1977perplexity}, CLIP-Score~\citep{hessel2021clipscore}, MM-Selector~\citep{ma2025mllm}, PRISM~\citep{bi2025prism}, ICONS~\citep{wu2024icons}, DSIR~\citep{xie2023data}, and DataTailor~\citep{yu2025mastering}. For \emph{group-level selection}, we choose COINCIDE~\citep{lee2024concept}, SemDeDup~\citep{abbas2021semdedup}, CLIMB~\citep{diao2025climb}, and RegMix~\citep{liu2024regmix}. Following their mixture-prediction strategies, CLIMB predicts target accuracy from semantic-cluster shares, while RegMix predicts proxy loss from source shares. For fair comparison, all selectors use the same candidate pool, 100K budget, and final-training recipe; probe and proxy settings are detailed in \Cref{app:baselines}. Random samples uniformly; Full uses all 8M. Because ICONS and DSIR require validation or reference data, both receive CLIVO's selected validation examples for a fair comparison with common reference content. CLIVO samples randomly within clusters by default. To test compatibility with sample ranking, CLIVO + ICONS applies the strong ICONS selector within unchanged quotas. Random, CLIMB, RegMix, and CLIVO report three-run means; others use one run (\Cref{app:reported-runs}).

\paragraph{Implementation details.}
We embed instructions with Qwen3-VL-Embedding-8B~\citep{li2026qwen3}, use $K=100$ clusters, and train experts with the nominal step-10 budget. Validation learning uses $Y=128$ real probes and $H=10{,}000$ candidates sampled equally across clusters, retaining $m=64$ examples per target and fitting their weights with $\lambda=10$. We run 120 gradient updates, retain the best periodically evaluated mixture, and train its 100K-example subset for 100 optimizer steps. Budget definitions, optimizer settings, and evaluation protocols are in \Cref{app:implementation,app:experimental-protocols}.

\WFclear
\Needspace*{50pt}
\subsection{Target-Specific Selection Quality}
\label{sec:main-results}

\Needspace*{225pt}
\setlength{\intextsep}{0pt}
\leavevmode\begin{wraptable}{r}{0.59\textwidth}
\centering\footnotesize
\setlength{\tabcolsep}{2pt}
\captionsetup{font=footnotesize,skip=4pt}
\begin{threeparttable}
\caption{\textbf{Selection quality at a 100K budget.} O/G/P/K: four targets; Average: their mean. Random, CLIMB, RegMix, and CLIVO average three runs. Bold/underline: best/second-best at 100K.}
\label{tab:main-results}
\begin{tabular*}{\linewidth}{@{\extracolsep{\fill}}lccccc@{}}
\toprule
Method & O & G & P & K & Average\\
\midrule
Full (8M) & 71.01 & 69.23 & 80.87 & 63.81 & 71.23\\
\midrule
Random & 66.61 & 63.57 & 77.73 & 60.05 & 66.99\\
PPL & 64.26 & 60.78 & 75.65 & 58.09 & 64.70\\
CLIP-Score & 66.16 & 62.23 & 79.42 & 58.80 & 66.65\\
MM-Selector & 66.46 & 64.05 & 76.96 & 59.59 & 66.77\\
PRISM & 67.27 & 65.73 & 76.97 & 59.86 & 67.46\\
ICONS & 67.32 & 64.12 & 79.30 & 59.88 & 67.65\\
DSIR & 67.31 & \underline{67.56} & 72.62 & 60.42 & 66.98\\
DataTailor & 65.72 & 61.50 & 77.58 & 60.20 & 66.25\\
SemDeDup & 64.00 & 57.53 & 77.87 & 59.84 & 64.81\\
COINCIDE & 66.20 & 62.65 & 78.14 & 59.60 & 66.65\\
CLIMB & 67.53 & 65.44 & 77.80 & 61.40 & 68.04\\
RegMix & 67.00 & 62.91 & 79.12 & 59.80 & 67.21\\
\midrule
CLIVO & \textbf{68.13} & 67.28 & \underline{79.64} & \underline{61.76} & \underline{69.20}\\
CLIVO + ICONS & \underline{68.05} & \textbf{68.27} & \textbf{79.96} & \textbf{62.09} & \textbf{69.59}\\
\bottomrule
\end{tabular*}
\end{threeparttable}
\end{wraptable}
\noindent\textbf{CLIVO achieves competitive fixed-budget selection.}
With random within-cluster sampling, CLIVO reaches a four-target mean of 69.20, exceeding CLIMB by 1.16 points and the strongest sample-level selector, ICONS, by 1.55 points (\Cref{tab:main-results}). It also surpasses RegMix by 1.99 points. CLIVO leads the baselines on Overall, Parsing, and Knowledge with scores of 68.13, 79.64, and 61.76, showing the value of target-specific semantic allocation without ranking individual training examples. Baselines favor different capabilities: CLIP-Score, which measures image--text alignment, achieves the highest baseline Parsing score (79.42); DSIR, which matches a reference distribution, achieves the highest baseline General score (67.56). ICONS and DSIR also obtain competitive results using CLIVO's selected validation examples, showing that this reference data can support existing selectors. Benchmark-level results are in \Cref{app:benchmark-exhibits}.

\noindent\textbf{Allocation and ranking are complementary.}
Keeping CLIVO's quotas fixed and replacing random sampling with within-cluster ICONS improves most targets, raising the four-target mean from 69.20 to 69.59 and General by 0.99 points. The combination also exceeds global ICONS on all four targets, with gains of 4.15 points on General and 1.94 on the four-target mean. Thus, the same ranking method is more effective when it fills target-specific quotas: allocation determines how much of each content type to include, and ranking chooses its representatives. Global sample ranking alone leaves these composition choices implicit (\Cref{app:within-selection}).

\WFclear
\Needspace*{60pt}
\subsection{Ablation Studies}
\label{sec:component-results}
\label{sec:sensor-evaluation}

We examine three questions: whether finer allocation improves selection,
whether proxy losses provide reliable target feedback, and whether
gradient search uses that feedback to improve mixtures.

\noindent\textbf{The value of finer allocation depends on the target.}
We isolate allocation freedom by tying the same 100 expert coefficients into 20 or 50 parent shares, or optimizing all 100 independently. The proxy bank, validation objective, starting composition, and training budgets stay fixed. Tying is algebraically equivalent to premerging child experts at fixed ratios (\Cref{app:unit-control}).

\begin{figure}[!htbp]
\centering
\includegraphics[width=\linewidth]{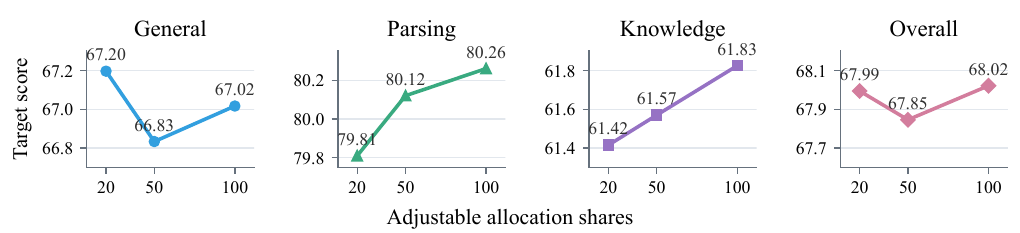}
\caption{\textbf{Allocation freedom and downstream performance.} Each panel varies adjustable shares on the same 100-expert bank. Scores use 100K examples, random within-cluster sampling, and one training seed. All vertical axes span 0.6 points. Search records: \Cref{tab:component-results}.}
\label{fig:allocation-freedom}
\end{figure}

In \Cref{fig:allocation-freedom}, 100 adjustable shares give the highest scores on most targets. Parsing and Knowledge improve by 0.45 and 0.41 points over 20 shares, whereas General performs best at 20. Finer allocation is useful when a target benefits from separating content within coarse groups; the preferred granularity therefore depends on the target.

\begin{figure}[!htbp]
\centering
\includegraphics[width=\linewidth]{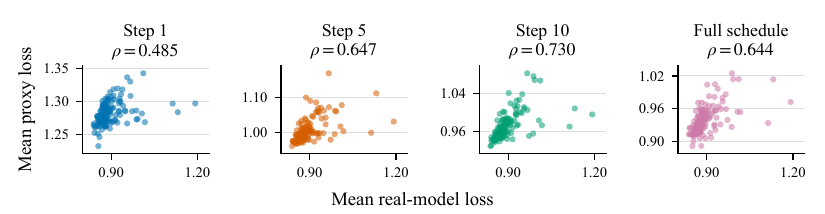}
\caption{\textbf{Expert budget and proxy--real loss agreement.} Each point compares mean proxy and real-model losses for the same mixture on a shared 10,000-example panel. Panels show different expert budgets; $\rho$ is Spearman correlation across 128 mixtures. Budget details: \Cref{app:expert-budget}.}
\label{fig:proxy-fidelity}
\end{figure}
\noindent\textbf{Moderate-budget experts capture real-model loss variation.}
To test proxy reliability, we compare expert checkpoints indexed by $b\in\{1,5,10\}$ and a full-training reference on identical mixtures and examples (\Cref{fig:proxy-fidelity}). Proxy--real loss correlation increases from 0.485 to 0.730 across the short budgets, with checkpoint 10 giving the strongest agreement. Full training lowers absolute loss but does not improve this agreement. Checkpoint 10 also improves held-out downstream-score prediction over checkpoint 1 on every target (\Cref{tab:expert-budget}). These results support its use as the shared proxy bank; budget definitions and protocols are in \Cref{app:expert-budget}.

\WFclear
\Needspace*{145pt}
\begin{wraptable}{r}{0.43\textwidth}
\centering\footnotesize
\captionsetup{font=footnotesize,skip=3pt}
\setlength{\tabcolsep}{2pt}
\begin{threeparttable}
\caption{\textbf{Validation aggregation.} Held-out Spearman over 20 folds, with 64 matched examples. C/S: cluster/sample; E/W: equal/weighted. Maximum correlations are bold.}
\label{tab:sensor-quality}
\begin{tabular*}{\linewidth}{@{\extracolsep{\fill}}lrrrr@{}}
\toprule
Target & C--E & C--W & S--E & S--W\\
\midrule
General & 0.621 & 0.684 & 0.619 & \textbf{0.714}\\
Parsing & 0.397 & 0.430 & 0.379 & \textbf{0.461}\\
Knowledge & 0.586 & 0.594 & 0.589 & \textbf{0.596}\\
Overall & 0.469 & 0.569 & 0.470 & \textbf{0.591}\\
\bottomrule
\end{tabular*}
\end{threeparttable}
\end{wraptable}
\noindent\textbf{Individual responses preserve useful validation information.}
We compare cluster-level and sample-level aggregation, each with equal or learned weights, on the same 64 examples per fitting fold (\Cref{tab:sensor-quality}). Cluster-level rules average losses within each represented cluster before weighting, while sample-level rules retain each example's response to mixture changes. Learned sample weights give the highest held-out Spearman on all four targets, improving over equal sample averaging by up to 0.121. Their advantage over weighted cluster means supports retaining individual responses: fitting can emphasize informative examples whose signals would otherwise be diluted by averaging. The fit concentrates weight on a subset of examples (\Cref{app:objective-construction}), emphasizing their contribution to target evaluation.

\WFclear
\Needspace*{155pt}
\begin{wraptable}{r}{0.59\textwidth}
\centering\footnotesize
\captionsetup{font=footnotesize,skip=3pt}
\setlength{\tabcolsep}{2pt}
\begin{threeparttable}
\caption{\textbf{From feedback to final training.} $K=100$. CV-$\rho$: held-out Spearman; $t^\star$: best step; $D_J$: loss decrease (\%); $B$: best probe; $\bar F$: three-run mean; $\Delta F=\bar F-B$. Protocols: \Cref{app:search-exhibits}.}
\label{tab:workflow-summary}
\begin{tabular*}{\linewidth}{@{\extracolsep{\fill}}lrrrrrr@{}}
\toprule
Target & CV-$\rho$ & $t^\star$ & $D_J$ & $B$ & $\bar F$ & $\Delta F$\\
\midrule
Gen. & 0.714 & 100 & 3.724 & 66.600 & 67.28 & $+0.68$\\
Pars. & 0.461 & 120 & 0.293 & 79.310 & 79.64 & $+0.33$\\
Know. & 0.596 & 40 & 0.056 & 61.465 & 61.76 & $+0.30$\\
Ovrl. & 0.591 & 100 & 2.040 & 67.919 & 68.13 & $+0.21$\\
\bottomrule
\end{tabular*}
\end{threeparttable}
\end{wraptable}
\noindent\textbf{Learned signals guide mixture search.}
Held-out correlations show that the learned objectives predict downstream rankings for all four targets (\Cref{tab:workflow-summary}). Gradient search lowers every fixed objective, reducing loss by 0.056--3.724\% from the evaluated starts. These signals therefore support direct optimization of the mixture shares. All four final-training means also exceed the best historical probe scores, by 0.21--0.68 points. \Cref{app:search-exhibits} provides search trajectories and training details.

\WFclear
\setlength{\intextsep}{8pt plus 2pt minus 2pt}

\section{Interpretable, Flexible, and Reusable Data Curation}
\label{sec:framework-analysis}

In this section, we show how CLIVO supports \textbf{interpretable, flexible, and reusable data curation}. Semantic visualizations make target-specific selection decisions inspectable (\Cref{sec:selected-data}); modular components accommodate different applications, while shared experts and feedback support reuse across targets and allocation granularities (\Cref{sec:reuse-extensibility}).

\begin{figure}[!htb]
\centering
\includegraphics[width=\linewidth]{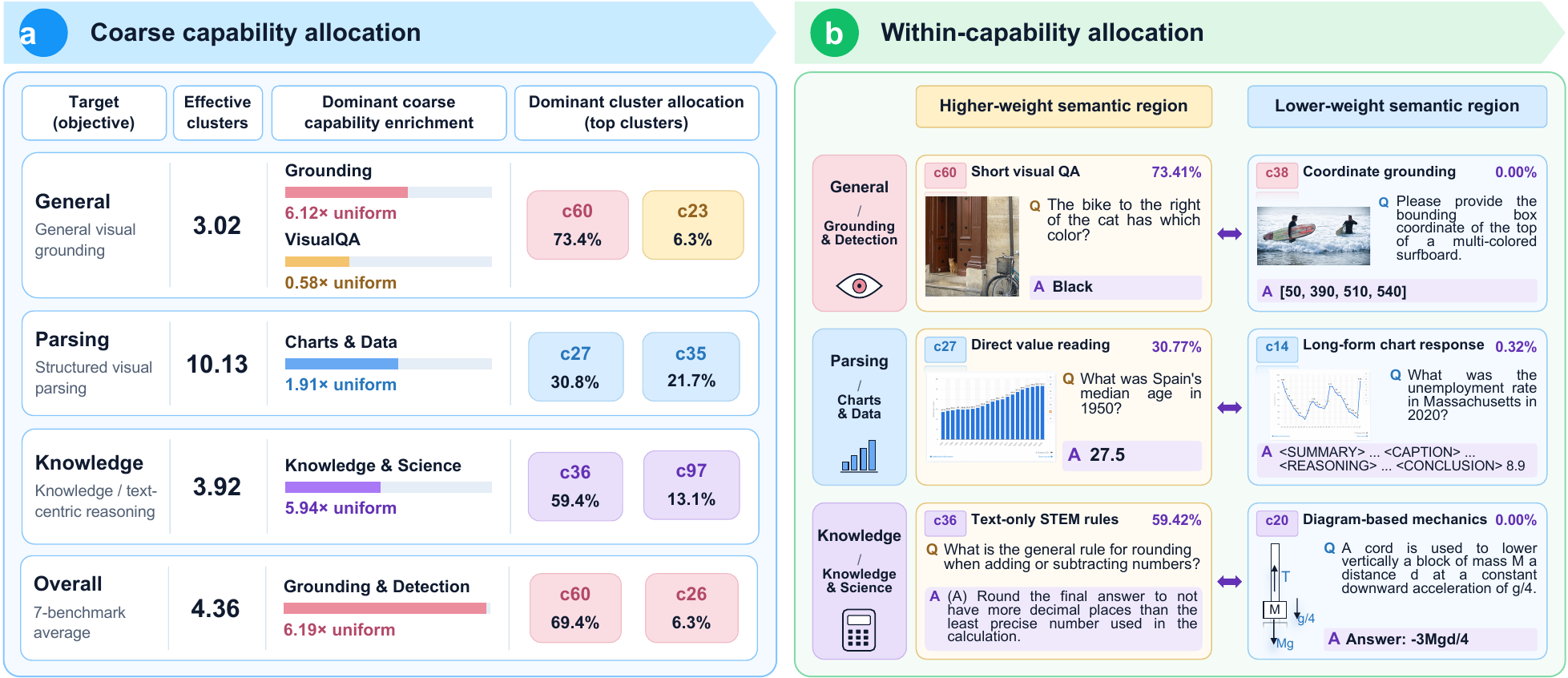}
\caption{\textbf{Target-specific composition, from distributions to examples.} \textbf{(a)} Allocation concentration and dominant semantic regions. \textbf{(b)} Paired clusters within the same coarse capability receive different training shares. Percentages give each cluster's share of the complete training budget. Uniform allocates 1\% to each of the 100 clusters; $\times$ uniform compares a category's total learned share with its share under this equal-cluster allocation. Definitions and cases: \Cref{app:data-analysis}.}
\label{fig:selected-data}
\end{figure}

\subsection{Data Visualization and Management}
\label{sec:selected-data}

\paragraph{Targets favor different, concentrated mixtures.}
We use GPT-based task annotations to group clusters into coarse capability categories for interpretation (\Cref{app:capability-annotations}). \Cref{fig:selected-data} shows distinct preferences: General and Overall emphasize c60, illustrated by short visual questions; Parsing favors c27, illustrated by chart reading; and Knowledge favors c36, illustrated by text-based STEM supervision. Allocation is concentrated but target-dependent: the entropy-based effective cluster count is 10.13 for Parsing and 3.02--4.36 for the other targets, out of 100 clusters. Labels and examples help interpret these differences (\Cref{tab:data-selection-diagnostics}).

\paragraph{Within-category contrasts highlight the value of fine-grained allocation.}
Clusters within the same coarse capability receive markedly different shares (\Cref{fig:selected-data}(b)). Within \emph{Grounding \& Detection}, General assigns 73.41\% to c60, illustrated by short visual questions, and zero to c38, illustrated by bounding-box coordinates. Within \emph{Charts \& Data}, Parsing assigns 30.77\% to c27, illustrated by direct value reading, and 0.32\% to c14, illustrated by long responses. Within \emph{Knowledge \& Science}, Knowledge assigns 59.42\% to c36, illustrated by text-only STEM tasks, and zero to c20, illustrated by mechanics diagrams. These contrasts reveal target preferences within broad categories, helping explain why independently adjustable fine-grained shares can be useful (\Cref{fig:allocation-freedom}).

\subsection{Flexible and Reusable Data Curation}
\label{sec:reuse-extensibility}

\paragraph{Modular components enable flexible selection.}
CLIVO separates semantic organization, target evaluation, and within-cluster
selection, so each component can be configured for the application.
Its quotas also integrate with state-of-the-art sample-level selectors:
allocation determines how much of each content type to include, and
ranking chooses the examples within those quotas. The CLIVO + ICONS
results demonstrate this combination (\Cref{sec:main-results}), allowing
existing ranking methods to benefit from target-specific composition.

\paragraph{One-time construction supports new targets.}
After constructing the embeddings, semantic partition, expert bank,
real probe models, and candidate-loss cache, CLIVO can reuse these assets
for new targets on the same base and partition. Evaluating the saved
probes on a new benchmark provides the target scores; validation-example
selection and weighting then operate on cached losses, followed by
gradient search under the new objective. Thus, target adaptation requires
neither retraining experts or probes nor recomputing the cached
per-example losses (\Cref{app:search-complexity}).

\paragraph{Shared gradients support finer allocation.}
As allocation becomes finer on the same expert bank, a single backward
pass supplies gradients to update all mixture weights jointly
(Eq.~\eqref{eq:mixture-gradient}). CLIVO thus avoids fitting an additional
regression model over the higher-dimensional mixture space. In the
$K=20/50/100$ experiments, costs to reach the best recorded losses show no
consistent monotonic increase across targets as allocation becomes finer
(\Cref{tab:component-results}). With the same expert bank and validation
objectives, the four gradient searches use 26.6\% fewer
peak-BF16-normalized GPU-hours than labeling 4,096 mixtures for regression,
a budget at which held-out proxy-loss prediction shows diminishing gains.
This reuse of model gradients
supports fine-grained search without an additional mixture-response
fitting stage (\Cref{app:search-complexity,app:incremental-cost}).

\section{Related Work}
\label{sec:related}

\paragraph{Sample-level methods.}
Sample-level methods score individual examples and select a subset according to those scores. Their signals fall into four broad categories. \emph{Quality and alignment} scores assess instruction quality and complexity in DEITA~\citep{liu2023makes}, human preferences in Align2LLaVA~\citep{huang2025align2llava}, and image--text compatibility in CLIP-Score~\citep{hessel2021clipscore}. \emph{Model-response} signals capture necessity in MLLM-Selector and selector confidence in Once-For-All~\citep{dong2026once}. \emph{Influence-based} methods estimate training effects: TIVE combines instance influence with task difficulty, ICONS seeks cross-task consensus, and BIDS~\citep{dai2025improving} balances task-dependent influence scales in text instruction tuning. \emph{Distribution and diversity} signals address coverage: DSIR~\citep{xie2023data} resamples toward a reference distribution, PRISM~\citep{bi2025prism} uses visual-feature geometry, DataTailor~\citep{yu2025mastering} combines informativeness, uniqueness, and representativeness, and PreSel~\citep{safaei2025filter} clusters images before generating instructions. CoIDO jointly models importance and diversity. These categories can overlap with allocation; TIVE and DataTailor also determine task quotas. CLIVO learns target-specific semantic quotas that can be filled using existing sample-level selectors.

\paragraph{Group-level methods.}
Group-level methods partition a pool and determine how much training data to draw from each group. They can be classified by how shares are chosen. \emph{Heuristic allocation} converts group statistics into quotas, as in COINCIDE's use of concept--skill density and transferability. \emph{Online adaptive weighting} updates shares during training; DoReMi uses group distributionally robust optimization on reference-relative losses. \emph{Response-model-based search} fits mixture outcomes, using flexible regressors in RegMix and CLIMB or parametric laws in Data Mixing Laws~\citep{ye2024data}, BiMix~\citep{ge2024bimix}, and CAMEL. The latter additionally links validation losses to benchmark scores. Expert reuse changes how these response models obtain feedback: MDE~\citep{belenki2025optimizing} constructs features from combined expert predictions, while DeMix and MergeMix evaluate parameter-merged experts to train LightGBM predictors~\citep{ke2017lightgbm}. \emph{Direct proxy evaluation} compares merged candidates without a fitted mixture-response function~\citep{berasi2026linear}. Related work on adaptive merging optimizes coefficients using entropy minimization to improve the merged model itself~\citep{yang2024adamerging}. CLIVO learns a sample-loss objective and differentiates it through fixed experts to optimize training shares.

\section{Conclusion}
\label{sec:conclusion}

We presented CLIVO, a target-aware instruction-selection framework connecting fine-grained semantic allocation, learned validation objectives, and differentiable mixture search. Learning from individual validation losses turns real-training feedback into an objective for optimizing data composition. Experiments show that learned quotas improve fixed-budget selection and complement sample ranking. This distinction between allocating content and ranking examples provides a practical way to adapt a shared instruction pool to different capability needs. The learned objective links these allocation decisions to target performance, while semantic visualization reveals which content is emphasized. Modular components accommodate different encoders and selectors, and shared experts support reuse across targets and granularities. Together, these properties make the selected subset both adaptable and open to inspection. Future work will explore adaptive probe selection to reduce real-training feedback, transfer of learned validation objectives across base models and training budgets, and joint optimization of multiple capability targets. Hierarchical semantic units could also adapt allocation granularity to the task as instruction pools grow, extending the framework to evolving data collections. We hope this work encourages joint study of example quality and data composition for transparent, target-aware curation.

\clearpage
\subsection*{AI Use Statement}
AI tools were used to polish the manuscript's language and improve readability.
GPT also assisted semantic annotation for data visualization; the annotation
procedure and human verification are described in \Cref{app:data-analysis}.
The authors independently developed the core research ideas and methodology,
designed and conducted the experiments, and formulated the scientific conclusions.
The authors take full responsibility for this work.

\subsection*{Ethics Statement}
This work uses publicly available datasets and derived instruction data in accordance with their applicable licenses and terms of use. We document the data sources and processing scope in \Cref{app:data-sources}. Our experiments do not involve recruiting human participants or collecting new personal information. Data processing and reporting follow applicable privacy requirements, and qualitative examples are presented without personally identifying information. CLIVO is intended to support transparent, target-aware data selection for research.

\subsection*{Reproducibility Statement}
\Cref{sec:method} specifies validation-example screening, nonnegative ridge
fitting, weight conversion, expert merging, and mixture search.
\Cref{app:theory} gives the method derivations and validation objective;
\Cref{app:implementation} documents data, training, baselines, and detailed
results. Annotation procedures, example provenance, and reuse accounting
appear in \Cref{app:framework-details}.

\bibliography{main/reference}
\bibliographystyle{iclr2027_conference}

\clearpage
\appendix

\section*{Appendix Contents}
\begingroup
\small
\setlength{\parskip}{2pt}
\newcommand{\appentry}[2]{\noindent\hspace*{#1}\hyperref[#2]{\ref*{#2}\quad\nameref*{#2}}\dotfill\pageref*{#2}\par}
\appentry{0pt}{app:theory}
\appentry{10pt}{app:unit-theory}
\appentry{10pt}{app:random-probes}
\appentry{10pt}{app:objective-construction}
\appentry{10pt}{app:decision-theory}
\appentry{0pt}{app:implementation}
\appentry{10pt}{app:model-construction}
\appentry{10pt}{app:data-sources}
\appentry{10pt}{app:training-evaluation}
\appentry{10pt}{app:baselines}
\appentry{10pt}{app:reported-runs}
\appentry{10pt}{app:benchmark-exhibits}
\appentry{10pt}{app:unit-control}
\appentry{10pt}{app:expert-budget}
\appentry{10pt}{app:search-exhibits}
\appentry{0pt}{app:framework-details}
\appentry{10pt}{app:data-analysis}
\appentry{10pt}{app:within-selection}
\appentry{10pt}{app:search-complexity}
\appentry{10pt}{app:incremental-cost}
\endgroup

\section{Method Details}
\label{app:theory}

\subsection{Finer Allocation Can Improve the Attainable Optimum}
\label{app:unit-theory}

A coarse allocation fixes the proportions of fine clusters within each
parent. Write this restriction as $\mathbf w=B\mathbf z$, where each
column of $B$ contains the nonnegative child proportions of one parent
and sums to one. The fine family $\mathcal W$ and its coarse restriction
$\mathcal Z=\{\mathbf z:B\mathbf z\in\mathcal W\}$ satisfy
\begin{equation}
\sup_{\mathbf z\in\mathcal Z}F_d(B\mathbf z)
\;\leq\;
\sup_{\mathbf w\in\mathcal W}F_d(\mathbf w).
\label{eq:unit-inclusion}
\end{equation}
Every coarse mixture is also a fine mixture, so optimizing over the
larger family cannot lower the best attainable score. The improvement
can be strict when a better composition requires changing the fixed
within-parent proportions. This is the benefit of fine-grained
allocation: it preserves coarse choices and allows additional ones.

\subsection{Mixture Interpretation of Within-Cluster Sampling}
\label{app:random-probes}

With uniform within-cluster sampling fixed, composition is the selection
decision being varied. For cluster size $M_k$, draw $n_k$ examples uniformly
without replacement and let $r_k=n_k/N$ be its realized share. For any
fixed per-example quantity $f$,
\begin{equation}
\mathbb E\!\left[\frac1N\sum_{x\in\mathcal S}f(x)\right]
=\sum_k r_k\left(\frac1{M_k}\sum_{x\in\mathcal D_k}f(x)\right).
\label{eq:random-mixture-mean}
\end{equation}
Indeed, each example in cluster $k$ is included with probability
$n_k/M_k$; summing its expected contribution gives the identity.
Thus the subset represents the specified mixture in expectation,
without introducing a preference for particular examples within a
cluster. Under the fixed training recipe, a real probe supplies a noisy
observation of that mixture's expected downstream score. Repeated
sampling reduces variation in the mean under the same within-cluster rule.

\subsection{Learning the Validation Objective}
\label{app:objective-construction}

The candidate panel contains 100 examples from each of 100 clusters.
We evaluate proxies at the same 128 mixtures as the real probes.
Each loss averages cross-entropy over that example's answer tokens,
so response length is normalized before examples are aggregated.

\paragraph{Screening and fitting.}
Let $X_{ij}=-\ell(\widetilde\theta(\mathbf w_i);v_j)$.
On fitting indices $\mathcal I$, the screening score is
\begin{equation}
q_{d,j}=\tfrac12\rho_{\rm Sp}(X_{\mathcal I,j},\mathbf y_{\mathcal I}^{(d)})
+\tfrac12\rho_{\rm Sp}(X_{\mathcal I_d^+,j},\mathbf y_{\mathcal I_d^+}^{(d)}),
\label{eq:sample-screen}
\end{equation}
where $\mathcal I_d^+$ is the higher-scoring half of the fitting probes.
We retain the top $m=64$ examples and standardize their negative losses
using fitting-probe means $\mu_j$ and standard deviations $s_j$.
We then fit their coefficients by Eq.~\eqref{eq:nonnegative-ridge},
with $\lambda=10$.
The fit uses centered numerical target scores; Spearman correlation
measures the resulting ordering.

\paragraph{Returning to loss units.}
\label{app:weight-conversion-proof}
With $c_j=\widehat\beta_j/s_j$, $C_d=\sum_jc_j>0$, and
$\alpha_{d,j}=c_j/C_d$, the fitted score is
\begin{equation}
\widehat y_i
=\bar y_d+\sum_j\widehat\beta_j\frac{X_{ij}-\mu_j}{s_j}
=b_d-C_dJ_d(\mathbf w_i),
\qquad b_d=\bar y_d-\sum_jc_j\mu_j .
\label{eq:score-loss-affine}
\end{equation}
Consequently, minimizing the normalized weighted loss preserves the
fitted score ranking. The final objective is fitted on all 128 probes
and frozen for search.

\paragraph{Preserving individual responses.}
\label{app:aggregation-theory}
A weighted average of cluster-mean losses assigns the same effective
weight to every selected example within a cluster. Individual weights
include this special case but also distinguish examples with different
responses to the training mixture. They therefore preserve information
that a cluster mean removes before fitting.

\paragraph{Matched evaluation.}
The four aggregation rules in \Cref{tab:sensor-quality} share the same 64 screened
examples in each of five repeated four-fold splits. Screening,
standardization, and weight fitting use only the 96 fitting probes;
the remaining 32 evaluate ranking. Cluster-level rules average losses
within each represented cluster; sample-level rules retain individual
losses. Equal weighting uses a uniform average; learned weighting uses
the same nonnegative ridge fit. The sample-weighted fits retain,
on average, 15.3/12.2/13.6/16.1 positive weights for
General/Parsing/Knowledge/Overall, respectively. All splits reuse the
same 128 development probes.

\subsection{Direct Search Avoids an Additional Response-Fitting Error}
\label{app:decision-theory}

Suppose a response regressor $h$ approximates the fixed validation
objective $J_d$ on a candidate family $\mathcal C$, with
$\sup_{\mathcal C}|h-J_d|\leq\xi$. If its selected mixture
$\widehat{\mathbf w}_h$ is $\eta$-optimal for $h$, then
\begin{equation}
J_d(\widehat{\mathbf w}_h)-\min_{\mathcal C}J_d
\leq \eta+2\xi .
\label{eq:response-fit-error}
\end{equation}
To see this, let $\mathbf w_J$ minimize $J_d$. Then
$J_d(\widehat{\mathbf w}_h)\leq h(\widehat{\mathbf w}_h)+\xi
\leq h(\mathbf w_J)+\eta+\xi
\leq J_d(\mathbf w_J)+\eta+2\xi$.
An $\eta$-optimal solution to $J_d$ has a gap of at most $\eta$,
avoiding the additional $2\xi$ term from regression.

For the downstream interpretation, assume
$|F_d(\mathbf w)-(a_d-c_dJ_d(\mathbf w))|\leq\epsilon_d$
on $\mathcal C$, with $c_d>0$. Direct $\eta$-optimal search then gives
$\max_{\mathcal C}F_d-F_d(\widehat{\mathbf w})
\leq2\epsilon_d+c_d\eta$, while the response-fit route gives
$2\epsilon_d+c_d(\eta+2\xi)$. Direct search uses the same signal
without the extra regression term.

\section{Experimental Details and Additional Results}
\label{app:implementation}
\label{app:experimental-protocols}
\label{app:complete-results}

\subsection{Model Construction}
\label{app:model-construction}

We retain the pretrained Qwen language backbone and randomly initialize
the visual head and alignment MLP. Alignment jointly trains the visual
head and MLP while keeping the language backbone fixed, using
\href{https://huggingface.co/datasets/liuhaotian/LLaVA-Pretrain}{LLaVA-Pretrain (558K)}.
We then perform multimodal mid-training on
\href{https://huggingface.co/datasets/mvp-lab/LLaVA-OneVision-1.5-Mid-Training-85M}{LLaVA-OneVision-1.5 Mid-Training-85M}~\citep{an2025llava}
to obtain the common base $\theta_0$.

\subsection{Data Pool}
\label{app:data-sources}

Some public datasets combine multiple constituent sources. We manually
separate these into smaller source branches to preserve their identities,
then remove duplicate and noisy records. The resulting instruction
collection contains approximately 40M processed records across 30 branches.
\Cref{tab:pool-source-counts} lists the upstream releases and processed
counts. Upstream sizes refer to the indicated release or branch; processed
rows follow our training format, which can combine multiple turns into
one conversation. Derived branches can therefore differ in counting unit
from their parent releases. Sampling 20\% from each semantic region gives the shared
8,007,947-example candidate pool. Every selection method draws its 100K
subset from this pool; Full uses the complete candidate pool.

\begin{table}[!htbp]
\centering\footnotesize
\setlength{\tabcolsep}{4pt}
\begin{threeparttable}
\caption{\textbf{Instruction-pool sources.} Public release sizes and
processed source-branch rows before 20\% sampling. Derived branches
list parent-release sizes.}
\label{tab:pool-source-counts}
\begin{tabular}{@{}p{0.43\linewidth}rrr@{}}
\toprule
Dataset or branch & Link & Upstream size & Processed rows\\
\midrule
ECD-10K Images & \href{https://huggingface.co/datasets/ChartFoundation/ECD-10k-Images}{HF} & $\sim$321.5K & 321,544\\
Honey-Data-1M & \href{https://huggingface.co/datasets/Open-Bee/Honey-Data-1M}{HF} & 1,072,111 & 1,070,304\\
Llama-Nemotron-VLM v1, VQA-9 & \href{https://huggingface.co/datasets/nvidia/Llama-Nemotron-VLM-Dataset-v1}{HF} & 46,745 & 46,745\\
MGrounding-630K & \href{https://huggingface.co/datasets/Michael4933/MGrounding-630k}{HF} & $\sim$630K & 623,979\\
MM-IFInstruct-23K & \href{https://huggingface.co/datasets/ChrisDing1105/MMIF-23k}{HF} & 22,554 & 22,555\\
MM-IFInstruct-23K, second branch & \href{https://huggingface.co/datasets/ChrisDing1105/MMIF-23k}{HF} & 22,554 & 22,551\\
SPIQA, training split & \href{https://huggingface.co/datasets/google/spiqa}{HF} & 262,524 & 262,524\\
CMM-Math, image-bearing train & \href{https://huggingface.co/datasets/ecnu-icalk/cmm-math}{HF} & 6,272 & 5,004\\
DocReason25K & \href{https://huggingface.co/datasets/mPLUG/DocReason25K}{HF} & 25,877 & 25,741\\
MP-DocReason51K & \href{https://huggingface.co/datasets/mPLUG/MP-DocReason51K}{HF} & 51,754 & 49,959\\
Enigmata, generated snapshot & \href{https://huggingface.co/datasets/BytedTsinghua-SIA/Enigmata-Data/tree/95f6920c2428c762ba6a6a2038af7dbe15eb7c75}{HF} & 217,541 & 4,758\\
IQBench & \href{https://github.com/phuvinhnguyen/IQBench}{Code} & 500 & 500\\
LLaVA-OneVision Stage-2, video & \href{https://huggingface.co/datasets/lmms-lab/LLaVA-OneVision-Data}{HF} & $\sim$350K videos & $\sim$1,090,000\\
LLaVA-OneVision Stage-2, image & \href{https://huggingface.co/datasets/lmms-lab/LLaVA-OneVision-Data}{HF} & $\sim$2.7M & 2,633,205\\
LLaVA-OneVision Stage-2, text & \href{https://huggingface.co/datasets/lmms-lab/LLaVA-OneVision-Data}{HF} & $\sim$450K & 522,701\\
LLaVA-OneVision-1.5 Instruct & \href{https://huggingface.co/datasets/mvp-lab/LLaVA-OneVision-1.5-Instruct-Data}{HF} & $\sim$21.2M & 21,592,501\\
LLaVAR-2 Hybrid-Instruct & \href{https://huggingface.co/datasets/LLaVAR-2/Hybrid-Instruct}{HF} & $\sim$42K images & 43,660\\
LongAlign-10K & \href{https://huggingface.co/datasets/zai-org/LongAlign-10k}{HF} & 9,888 & 9,861\\
LongWriter-6K & \href{https://huggingface.co/datasets/zai-org/LongWriter-6k}{HF} & 6,000 & 5,926\\
Magpie-Align, filtered & \href{https://huggingface.co/collections/Magpie-Align/magpie-llama31-datasets}{HF} & 500K & 479,743\\
MMDU-45K & \href{https://huggingface.co/datasets/laolao77/MMDU}{HF} & 45,208 & 28,979\\
MM-MathInstruct-3M & \href{https://huggingface.co/datasets/MathLLMs/MM-MathInstruct}{HF} & 2,871,988 & 2,871,988\\
MTVQA, training split & \href{https://huggingface.co/datasets/ByteDance/MTVQA}{HF} & 6,678 images & 6,678\\
SmolTalk2, SFT & \href{https://huggingface.co/datasets/HuggingFaceTB/smoltalk2}{HF} & 3,383,242 & 3,382,846\\
MMPR-v1.2 & \href{https://huggingface.co/datasets/OpenGVLab/MMPR-v1.2}{HF} & 815,293 & 814,687\\
Tulu-3 Persona-IF & \href{https://huggingface.co/datasets/allenai/tulu-3-sft-personas-instruction-following}{HF} & 29,980 & 29,980\\
VisualWebInstruct, conversation & \href{https://huggingface.co/datasets/TIGER-Lab/VisualWebInstruct}{HF} & 1,004,070 & 1,004,068\\
VisualWebInstruct, text derivative & \href{https://huggingface.co/datasets/TIGER-Lab/VisualWebInstruct}{HF} & 1,004,070 & 558,467\\
COYO-derived instructions & \href{https://huggingface.co/datasets/kakaobrain/coyo-700m}{HF} & 746,972,269 & 99,965\\
WildChat branch & \href{https://huggingface.co/datasets/allenai/WildChat-4.8M}{HF} & 3,199,860 & 2,398,331\\
\bottomrule
\end{tabular}
\end{threeparttable}
\end{table}

\subsection{Training and Evaluation Settings}
\label{app:training-evaluation}

We report compute on a common 64-H100-equivalent basis.
Model training uses AdamW, BF16, and zero weight decay, with the schedules
in \Cref{tab:training-record}. Experts use a constant learning rate;
other stages use cosine decay with 1\% warmup. Final tuning uses a global
batch of 128 packed sequences for 100 updates. Training/loader seed pairs
are 42/52, 43/53, and 44/54; single-run comparisons use 42/52.

\begin{table}[!htbp]
\centering\small
\begin{threeparttable}
\caption{\textbf{Training schedules.} Expert checkpoint indices are
defined in \Cref{app:expert-budget}.}
\label{tab:training-record}
\begin{tabular}{@{}lrrr@{}}
\toprule
Stage & Updates / checkpoint & Peak learning rate & Token budget\\
\midrule
Alignment & 4,500 & $10^{-3}$ & 16,384\\
Mid-training & 35,999 of 40,000 & $10^{-5}$ & 8,192\\
Cluster experts & 1 / 5 / 10 & $10^{-5}$ & 16,384\\
Real probes & 50 & $10^{-5}$ & 16,384\\
Final instruction tuning & 100 & $10^{-5}$ & 16,384\\
\bottomrule
\end{tabular}
\end{threeparttable}
\end{table}

\paragraph{Selection settings.}
Search updates only the FP32 mixture shares using Adam at learning rate
$10^{-3}$ for 120 steps, with zero weight decay. Each update uses all
positive-weight validation examples; full BF16 evaluations every 20 steps,
including step 0, select the best mixture. Integer quotas total 100K by
rounding shares down and distributing the remainder to the largest
fractional parts, followed by uniform within-cluster sampling.

\paragraph{Target scores.}
For target benchmark set $\mathcal B_d$,
$y_i^{(d)}=|\mathcal B_d|^{-1}\sum_{b\in\mathcal B_d}s_{i,b}$.
Scores use percentage-point units, including OCRBench rescaled to
that range. Overall averages all seven benchmarks; Average in the
main table averages the four target columns.

\begin{table}[!htbp]
\centering\small
\begin{threeparttable}
\caption{\textbf{Benchmark-to-target mapping.} Domain targets average
their constituent benchmark scores equally. Overall averages all seven
scores, not the three domain means.}
\label{tab:benchmark-record}
\begin{tabular}{@{}llcc@{}}
\toprule
Benchmark & Domain target & Domain weight & Overall weight\\
\midrule
MMBench-EN & General & $1/3$ & $1/7$\\
MMStar & General & $1/3$ & $1/7$\\
GQA & General & $1/3$ & $1/7$\\
OCRBench & Parsing & $1/2$ & $1/7$\\
ChartQA & Parsing & $1/2$ & $1/7$\\
MMLU & Knowledge & $1/2$ & $1/7$\\
MMLU-Pro & Knowledge & $1/2$ & $1/7$\\
\bottomrule
\end{tabular}
\end{threeparttable}
\end{table}

\subsection{Baseline Selection Methods}
\label{app:baselines}

All selection methods construct 100K-example subsets from the same
candidate pool and use the same final-training recipe.
\textbf{Random} samples uniformly without replacement;
\textbf{Full} uses all 8M.

\paragraph{Sample-level methods.}
\textbf{PPL}~\citep{jelinek1977perplexity} favors examples with lower response perplexity,
using model likelihood as the selection signal.
\textbf{CLIP-Score}~\citep{hessel2021clipscore} uses image--text embedding similarity to favor
visually aligned instructions.
\textbf{MLLM-Selector}~\citep{ma2025mllm}, labeled MM-Selector in our tables,
combines necessity scores from a seed instruction-tuned model with
diversity-aware sampling.
\textbf{PRISM}~\citep{bi2025prism} derives training-free selection scores from intrinsic
visual features and their redundancy.
\textbf{ICONS}~\citep{wu2024icons} estimates gradient influence on reference examples and
aggregates task-level evidence by consensus.
\textbf{DSIR}~\citep{xie2023data} estimates source-to-reference importance ratios in a
lightweight feature space and resamples toward the reference
distribution.
\textbf{DataTailor}~\citep{yu2025mastering} combines informativeness, within-cluster uniqueness,
and cross-cluster representativeness.
Because ICONS and DSIR require validation or reference data, both receive
CLIVO's selected validation examples to keep target-specific reference
content consistent across comparisons.

\paragraph{Group-based methods.}
\textbf{SemDeDup}~\citep{abbas2021semdedup} identifies semantically similar examples in embedding
clusters and removes redundant representatives.
\textbf{COINCIDE}~\citep{lee2024concept} organizes examples by concept--skill representations
and selects using cluster density and transferability.
Following \textbf{CLIMB}~\citep{diao2025climb}, we use cluster-based
mixture prediction: a predictor maps semantic-mixture vectors to real
target scores, searches for high predicted accuracy, and fills the
selected cluster quotas uniformly. To control feedback cost, we fit
the predictor on the same fixed bank of 128 real probes without
additional bootstrapping rounds.
Following \textbf{RegMix}~\citep{liu2024regmix}, we use source-based
mixture regression over the 30 source branches. For consistent proxy
evaluation, we construct regression labels from target-aligned
merged-expert losses and sample uniformly within each selected source
quota. Both comparisons use the same candidate pool, 100K selection
budget, and final-training recipe as CLIVO.

\paragraph{Within-cluster variant.}
CLIVO uses uniform sampling by default. CLIVO + ICONS preserves its
learned quotas and selects the highest-scoring ICONS examples within
each cluster. Global ICONS instead selects across the entire pool.
Both use the same target-specific reference examples.

\subsection{Repeated Runs}
\label{app:reported-runs}

Random, CLIMB, RegMix, and CLIVO report three-run means; the other
methods report one run. CLIVO's repeats keep the selected mixture
fixed and vary the final sampling/training seed.
\Cref{tab:repeat-statistics} gives the mean, sample standard deviation,
and sample variance, using the $n-1$ denominator.
\Cref{tab:baseline-repeat-statistics} reports the available baseline
statistics: Random, CLIMB, and RegMix use population
standard deviation over three data-sampling seeds (42, 43, and 44).

\begin{table}[!htbp]
\centering\small
\begin{threeparttable}
\caption{\textbf{CLIVO final-training repeats.} Three runs at fixed searched
shares with uniform within-cluster sampling. Variance uses the $n-1$
denominator. }
\label{tab:repeat-statistics}
\begin{tabular}{@{}lrrr@{}}
\toprule
Target & Mean & Standard deviation & Sample variance\\
\midrule
General & 67.28 & 0.53 & 0.28\\
Parsing & 79.64 & 0.59 & 0.34\\
Knowledge & 61.76 & 0.11 & 0.01\\
Overall & 68.13 & 0.13 & 0.02\\
\bottomrule
\end{tabular}
\end{threeparttable}
\end{table}

\begin{table}[!htbp]
\centering\small
\begin{threeparttable}
\caption{\textbf{Baseline repeats.} Three-seed mean $\pm$ population
standard deviation.}
\label{tab:baseline-repeat-statistics}
\begin{tabular}{@{}lrcccc@{}}
\toprule
Method & Runs & Overall & General & Parsing & Knowledge\\
\midrule
Random & 3 & $66.61\pm0.23$ & $63.57\pm0.28$ & $77.73\pm0.09$ & $60.05\pm0.44$\\
RegMix & 3 & $67.00\pm0.17$ & $62.91\pm0.39$ & $79.12\pm0.26$ & $59.80\pm0.16$\\
CLIMB & 3 & $67.53\pm0.05$ & $65.44\pm0.38$ & $77.80\pm0.03$ & $61.40\pm0.26$\\
\bottomrule
\end{tabular}
\end{threeparttable}
\end{table}

\subsection{Benchmark-Level Results}
\label{app:benchmark-exhibits}

\Cref{tab:benchmark-results} reports CLIVO's three-run benchmark means
and the single-run CLIVO + ICONS results. Each target has its own
selected subset and trained model, so its benchmark profile is reported
separately. Domain scores and the four-target Average are calculated
before rounding.

\begin{table}[!htbp]
\centering\small
\setlength{\tabcolsep}{9pt}
\begin{threeparttable}
\caption{\textbf{Benchmark scores by target (\%).} CLIVO: three runs with
random sampling; CLIVO + ICONS: one run, same quotas.}
\label{tab:benchmark-results}
\begin{tabular}{@{}llrr@{}}
\toprule
Target & Benchmark & CLIVO & CLIVO + ICONS\\
\midrule
\multirow{3}{*}{General} & MMBench-EN & 84.87 & 85.22\\
 & MMStar & 55.98 & 58.27\\
 & GQA & 60.99 & 61.32\\
\addlinespace[3pt]
\multirow{2}{*}{Parsing} & OCRBench & 81.23 & 81.00\\
 & ChartQA & 78.04 & 78.92\\
\addlinespace[3pt]
\multirow{2}{*}{Knowledge} & MMLU & 72.23 & 72.25\\
 & MMLU-Pro & 51.29 & 51.93\\
\addlinespace[3pt]
\multirow{7}{*}{Overall} & MMBench-EN & 84.03 & 84.43\\
 & MMStar & 53.38 & 53.00\\
 & GQA & 59.87 & 60.68\\
 & OCRBench & 80.93 & 79.90\\
 & ChartQA & 77.01 & 77.28\\
 & MMLU & 70.74 & 70.71\\
 & MMLU-Pro & 50.93 & 50.38\\
\bottomrule
\end{tabular}
\end{threeparttable}
\end{table}

\subsection{Allocation Freedom on a Fixed Expert Bank}
\label{app:unit-control}

We keep the same 100 experts, initial mixture, validation objective,
120-update search, and final-training recipe. A nested Ward hierarchy
of cluster centroids groups the experts into 20 or 50 parents; using 100
shares leaves every expert's coefficient independently adjustable.
For parent $g$, fix child ratios
$a_{k|g}=w_k^0/\sum_{h\in\mathcal G_g}w_h^0$.
A zero-mass parent uses ratios proportional to child cluster sizes.
The optimized parent shares $\mathbf z$ expand as $\mathbf w=B\mathbf z$,
where $B_{kg}=a_{k|g}$ within parent $g$ and zero elsewhere.

\paragraph{Exact premerging identity.}
Define $\overline\Delta_g=\sum_{k\in\mathcal G_g}a_{k|g}\Delta_k$.
Then
\begin{equation}
\widetilde\theta(B\mathbf z)
=\theta_0+\sum_g z_g\overline\Delta_g .
\label{eq:tying-map}
\end{equation}
Thus tying fine coefficients is exactly equivalent to merging fixed
coarse experts $\overline\theta_g=\theta_0+\overline\Delta_g$.
The parent gradients are $B^\top\nabla_{\mathbf w}J_d$.
This isolates allocation freedom with fixed experts.

\paragraph{Relation to briefly training a coarse expert.}
Let $L_k$ be the mean training loss of child $k$ and
$L_g=\sum_ka_{k|g}L_k$.
For one gradient-descent step from the common $\theta_0$, with common
learning rate $\eta$,
\[
\sum_ka_{k|g}\bigl(\theta_0-\eta\nabla L_k(\theta_0)\bigr)
=\theta_0-\eta\nabla L_g(\theta_0).
\]
Thus, merging after one gradient-descent step exactly matches an update
on the coarse mixture. For smooth losses and a fixed number $T$ of small
gradient-descent steps,
$\theta_k^{(T)}=\theta_0-T\eta\nabla L_k(\theta_0)+O(\eta^2)$.
Averaging this expansion gives the coarse model's first-order update,
with a difference of order $O(\eta^2)$.
The local approximation motivates premerging. Our experiment uses the
exact identity in Eq.~\eqref{eq:tying-map}.

\begin{table}[!htbp]
\centering\small
\setlength{\tabcolsep}{6pt}
\begin{threeparttable}
\caption{\textbf{Allocation freedom.} $J^\star$: minimum proxy loss;
$t^\star$: best step; $C^\star$: 64-H100-equivalent hours; $F$: single-seed
score. Bold: best $F$. Accounting: \Cref{app:incremental-cost}.}
\label{tab:component-results}
\begin{tabular}{@{}lrrrrr@{}}
\toprule
Target & $K$ & $J^\star$ & $t^\star$ & $C^\star$ & $F$\\
\midrule
\multirow{3}{*}{General} & 20 & 0.32639 & 120 & 0.662 & \textbf{67.197}\\
 & 50 & 0.32801 & 80 & $\geq0.343$ & 66.833\\
 & 100 & 0.32145 & 100 & 0.808 & 67.017\\
\addlinespace[2pt]
\multirow{3}{*}{Parsing} & 20 & 0.85952 & 0 & 0.00 & 79.810\\
 & 50 & 0.85965 & 100 & 0.833 & 80.120\\
 & 100 & 0.85866 & 120 & 0.993 & \textbf{80.260}\\
\addlinespace[2pt]
\multirow{3}{*}{Knowledge} & 20 & 0.68731 & 40 & $\geq0.369$ & 61.415\\
 & 50 & 0.68743 & 120 & 1.128 & 61.570\\
 & 100 & 0.68760 & 40 & 0.402 & \textbf{61.825}\\
\addlinespace[2pt]
\multirow{3}{*}{Overall} & 20 & 0.37324 & 40 & 0.360 & 67.994\\
 & 50 & 0.37097 & 100 & 0.865 & 67.846\\
 & 100 & 0.36904 & 100 & 0.818 & \textbf{68.021}\\
\bottomrule
\end{tabular}
\end{threeparttable}
\end{table}

\Cref{tab:component-results} gives the full results underlying
\Cref{fig:allocation-freedom}. Each cell uses the same single-seed
training protocol. Parsing and Knowledge improve with finer shares;
General peaks at 20; Overall is not monotone.

\subsection{Expert Training Budget and Proxy Agreement}
\label{app:expert-budget}

We train one expert per cluster from the common base on 64 H100 GPUs
at a constant learning rate of $10^{-5}$, with four packed sequences
per GPU and a 16,384-token budget per sequence. We compare checkpoints
indexed $b=1,5,10$, holding this recipe fixed.
The step-10 source checkpoints record 11
completed updates and a median of 16,591 processed examples per expert.
A fully trained bank provides the reference. All banks use the same
128 mixtures and 10,000 examples.

\begin{table}[!htbp]
\centering\small
\setlength{\tabcolsep}{5pt}
\begin{threeparttable}
\caption{\textbf{Expert budget and two kinds of agreement.} Fidelity compares proxy with real-model loss over 128 mixtures. G/P/K/O are held-out correlations with real benchmark scores after budget-specific validation-cluster selection. Bold: highest in each correlation column.}
\label{tab:expert-budget}
\begin{tabular}{@{}lrrrrrr@{}}
\toprule
Bank & Mean loss & Fidelity $\rho$ & G $\rho$ & P $\rho$ & K $\rho$ & O $\rho$\\
\midrule
Step 1 & 1.283 & 0.485 & 0.383 & 0.435 & 0.523 & 0.367\\
Step 5 & 1.007 & 0.647 & 0.409 & \textbf{0.462} & 0.524 & 0.387\\
Step 10 & 0.972 & \textbf{0.730} & 0.518 & 0.448 & \textbf{0.527} & \textbf{0.460}\\
Full schedule & 0.940 & 0.644 & \textbf{0.577} & 0.382 & 0.440 & 0.457\\
\bottomrule
\end{tabular}
\end{threeparttable}
\end{table}

Fidelity is the Spearman correlation between proxy and real-model
mean losses over the 128 mixtures. For downstream prediction, five
repeated four-fold splits use 96 fitting and 32 held-out probes.
Within each fitting fold, clusters are ranked by the correlation between
negative mean loss and target score; three-fold inner validation chooses
the number retained from
$\{1,2,4,8,16,32\}$. The final columns report mean held-out correlations
after budget-specific selection. The strongest proxy--real agreement
and useful score prediction across targets support using step 10 as
the shared expert bank.

\subsection{Gradient Search Progress}
\label{app:search-exhibits}

\begin{figure}[!htb]
\centering
\includegraphics[width=\linewidth]{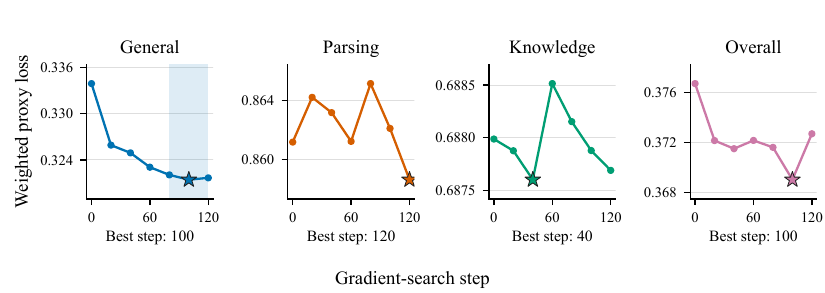}
\caption{\textbf{Search progress on fixed validation objectives.}
Stars mark the lowest recorded losses. General's shaded region retains
at least 95\% of its observed loss decrease with two subsequent checks.
Lines connect full-panel evaluations at 20-update intervals.}
\label{fig:search-trajectories}
\end{figure}

The four searches reduce their frozen validation objectives
(\Cref{fig:search-trajectories,tab:search-record}).
Full-panel evaluations every 20 updates select the best checkpoint,
at steps 100, 120, 40, and 100 for General, Parsing, Knowledge, and
Overall. General enters a near-best plateau at step 80, whereas losses for
the other targets continue to fluctuate. Periodic evaluation keeps
the best observed mixture throughout the 120-update search.

\begin{table}[!htbp]
\centering\scriptsize
\setlength{\tabcolsep}{4pt}
\begin{threeparttable}
\caption{\textbf{Search starts and checkpoints.} Full BF16 evaluation
selects $t^\star$; remerged step-zero loss $J_0$ defines the decrease $D_J$.
}
\label{tab:search-record}
\begin{tabular}{@{}lrrrr@{}}
\toprule
Target & Start recipe & $t^\star$ & $J_0$ & $J^\star$\\
\midrule
General & 101 & 100 & 0.333885 & 0.321450\\
Parsing & 119 & 120 & 0.861185 & 0.858662\\
Knowledge & 27 & 40 & 0.687988 & 0.687602\\
Overall & 101 & 100 & 0.376730 & 0.369045\\
\bottomrule
\end{tabular}
\end{threeparttable}
\end{table}

The decrease in \Cref{tab:workflow-summary} is
$D_J=100(J_0-J^\star)/J_0$, using the same run's evaluated start.
Best-of-probe references are 66.600, 79.310, 61.465, and 67.9186 for
General, Parsing, Knowledge, and Overall; the final score column uses
three-run means. \Cref{tab:training-record} lists the probe and final-model budgets.

\section{Data Interpretation, Flexibility, and Reuse}
\label{app:framework-details}

\subsection{Data Annotation and Composition}
\label{app:data-analysis}

We sample a 10,000-example panel across the 100 clusters for annotation
with \texttt{gpt-5.2-2025-12-11}; the archive contains 9,555 valid outputs.
The sample prompt labels task, content, format, and difficulty.
The cluster prompt summarizes label frequencies and representative
annotations. The response limits are 500 and 800 tokens, respectively;
images use a maximum side length of 512 pixels and videos supply up to
ten frames. An author-conducted check of sampled annotations yielded
a human--GPT agreement rate of 0.92. These semantic labels support data
visualization and interpretation. \Cref{tab:annotation-prompts} gives both prompts.

\begingroup
\small
\setlength{\LTcapwidth}{\linewidth}
\setlength{\LTpre}{6pt}
\setlength{\LTpost}{6pt}
\begin{longtable}{@{}p{\linewidth}@{}}
\caption{\textbf{GPT annotation prompts.} Sample labels describe task, content, format, and difficulty; cluster summaries aggregate these annotations.}
\label{tab:annotation-prompts}\\
\toprule
\endfirsthead
\caption[]{\textbf{GPT annotation prompts (continued).}}\\
\toprule
\endhead
\midrule
\multicolumn{1}{r}{\footnotesize Continued on the next page.}\\
\endfoot
\bottomrule
\endlastfoot
\textbf{Sample annotation.} The interaction's text and visual inputs follow this prompt.\\*
\begin{minipage}[t]{\linewidth}
\begin{lstlisting}[basicstyle=\ttfamily\fontsize{8.5}{9.5}\selectfont,aboveskip=2pt,belowskip=2pt,breaklines=true,columns=fullflexible,keepspaces=true,showstringspaces=false]
You are an expert Data Annotation Specialist. Your task is to analyze User-Assistant interaction samples and classify them into 4 dimensions using STRICT standardized labels.

The data comes from diverse multi-modal sources including VQA, Document Reasoning, Mathematical Logic, Web Agents, and Grounding tasks. You must output a SINGLE valid JSON object.

### 1. Task Dimension (Hierarchical)
Classify the user's primary intent.
- "task_category": Choose one of ["Perception", "Reasoning", "Generation", "Knowledge", "Chat"]
- "task_type": Choose the specific sub-type based on the category:
    - If Perception: ["OCR", "Object Detection", "Image Captioning", "Visual Grounding", "Video Tracking"]
    - If Reasoning: ["Math Solving", "Chart/Data Analysis", "Document Understanding", "Logic/Puzzle", "Web/UI Navigation", "Scientific Analysis"]
    - If Generation: ["Creative Writing", "Translation", "Summarization", "Code Implementation", "Long-form Content"]
    - If Knowledge: ["Fact Lookup", "Concept Explanation", "Trivia", "How-to Guide"]
    - If Chat: ["Roleplay", "General Chat", "Multi-turn Dialogue"]

### 2. Content Dimension (Visual & Semantic split)
Analyze the domain and visual inputs.
- "visual_content": Choose one: ["No Image", "Natural Scene", "Human/Face", "Document/Text", "Chart/Graph/Plot", "UI/Webpage", "Art/Illustration", "Abstract/Pattern"]
- "text_domain": Choose one: ["General/Daily Life", "STEM", "Humanities/Art", "Medical/Health", "Legal/Finance", "Code/Programming", "Web/Internet"]

### 3. Format Dimension (Input & Output Structure)
Analyze the structural form.
- "input_composition": Choose one: ["Text-Only", "Single-Image", "Multi-Image", "Video", "Interleaved-Text-Image"]
- "response_style": Choose one: ["Multiple Choice", "Yes/No", "Short Answer", "Long Explanation/CoT", "Code Block", "Structured Format (JSON/Table)", "Bounding Box/Coordinates"]

### 4. Difficulty Dimension (Cognitive Load)
- "difficulty_level": Choose one: ["Low", "Medium", "High"]

**CRITICAL:** Output raw JSON only. No markdown. No explanations.
\end{lstlisting}
\end{minipage}\\
\midrule

\textbf{Cluster summary.} Inputs follow under \texttt{Cluster Aggregated Statistics}
(label histograms) and \texttt{Sampled Annotations} (up to 100 records).\\*
\begin{minipage}[t]{\linewidth}
\begin{lstlisting}[basicstyle=\ttfamily\fontsize{8.5}{9.5}\selectfont,aboveskip=2pt,belowskip=2pt,breaklines=true,columns=fullflexible,keepspaces=true,showstringspaces=false]
You are an expert Data Analyst and Dataset Curator. 
I will provide you with the aggregated annotation statistics and a sample of individual interactions from a specific data cluster. These individual samples were previously annotated across 4 strict dimensions: Task, Content, Format, and Difficulty.

Your objective is to analyze this aggregated data and the samples to construct a comprehensive, multi-dimensional profile of this specific cluster.

Please output a valid JSON object strictly matching the following schema. Use the predefined label vocabularies from the sample annotations when identifying dominant categories.

{
    "task_dimension_summary": {
        "dominant_task_category": "The most frequent task category (e.g., Perception, Reasoning, etc.)",
        "top_task_types": ["List the top 1-3 specific task_types present in this cluster"],
        "task_diversity_note": "A 1-sentence analysis of whether this cluster is highly focused on one task or mixes several."
    },
    "content_dimension_summary": {
        "primary_visual_content": ["List the top 1-2 most frequent visual_content types"],
        "dominant_text_domain": "The most frequent text_domain",
        "content_fusion_analysis": "A 1-sentence description of how the visual and textual domains interact in this cluster."
    },
    "format_dimension_summary": {
        "dominant_input_composition": "The most frequent input_composition (e.g., Interleaved-Text-Image)",
        "typical_response_style": ["List the top 1-2 expected response_styles"]
    },
    "difficulty_dimension_summary": {
        "overall_difficulty_level": "Low / Medium / High / Mixed (based on majority distribution)",
        "cognitive_load_description": "A 1-sentence explanation of why the data requires this level of cognitive load."
    },
    "comprehensive_cluster_profile": {
        "short_definition": "A precise, 1-sentence title or definition for this cluster (e.g., 'Complex Scientific Document Reasoning and Formatting').",
        "detailed_description": "A comprehensive 3-4 sentence paragraph synthesizing all 4 dimensions. Describe the typical user intent, the nature of the multi-modal inputs, the expected output structure, and the overall technical or cognitive demands of the data within this cluster."
    }
}

**CRITICAL INSTRUCTIONS:**
1. Output raw JSON ONLY. 
2. Do NOT wrap the JSON in markdown code blocks (e.g., no ```json ... ```). 
3. Do NOT include any introductory or concluding text.
4. Ensure all dimensional analyses strictly reflect the provided sample data and statistics.
\end{lstlisting}
\end{minipage}\\
\end{longtable}
\endgroup

\paragraph{From fine tasks to coarse capabilities.}
\label{app:capability-annotations}
We map the first task in each cluster summary's
\texttt{top\_task\_types} to a coarse capability using
\Cref{tab:capability-taxonomy}. These annotations make the learned
allocation interpretable; embeddings define clusters and the target
objective sets their shares.

\begin{table}[!htbp]
\centering\small
\begin{threeparttable}
\caption{\textbf{Coarse capability labels for visualization.} Each cluster's
primary annotated task determines its coarse label. The original task
descriptions are retained for finer interpretation.}
\label{tab:capability-taxonomy}
\begin{tabular}{@{}lp{0.62\linewidth}@{}}
\toprule
Coarse capability & Primary task types\\
\midrule
Visual QA \& Captioning & Image Captioning\\
Grounding \& Detection & Object Detection; Visual Grounding; Video Tracking\\
OCR \& Documents & OCR; Document Understanding\\
Charts \& Data & Chart/Data Analysis\\
Math \& Logic & Math Solving; Logic/Puzzle\\
Knowledge \& Science & Scientific Analysis; Concept Explanation; Fact Lookup\\
Generation \& Other & Remaining task types, including code and creative generation\\
\bottomrule
\end{tabular}
\end{threeparttable}
\end{table}

\paragraph{Composition statistics and cases.}
For normalized shares, the effective cluster count is
$K_{\rm eff}=\exp(-\sum_{k:w_k>0}w_k\log w_k)$:
3.024, 10.134, 3.923, and 4.362 for General, Parsing, Knowledge, and
Overall. Uniform allocation gives each cluster 1\%; a category $A$
therefore has uniform share $|A|/100$ and enrichment
$(\sum_{k\in A}w_k)/(|A|/100)$.
The within-capability cases pair c60/c38 for General, c27/c14 for
Parsing, and c36/c20 for Knowledge. Percentages are shares of the total budget.

\Cref{tab:data-selection-diagnostics} compares the complete mixtures.
Total variation is $\tfrac12\sum_k|p_k-q_k|$; Jensen--Shannon divergence
is $\tfrac12\mathrm{KL}_2(p\|m)+\tfrac12\mathrm{KL}_2(q\|m)$,
where $m=(p+q)/2$.

\begin{table}[!htbp]
\centering\small
\begin{threeparttable}
\caption{\textbf{Target-dependent mixtures.} TV and JSD compare final
cluster shares.}
\label{tab:data-selection-diagnostics}
\begin{tabular}{@{}lcc@{}}
\toprule
Target pair & TV & JSD (bits)\\
\midrule
General / Parsing & 0.9783 & 0.9481\\
General / Knowledge & 0.9999 & 0.9995\\
General / Overall & 0.2610 & 0.2126\\
Parsing / Knowledge & 0.9895 & 0.9789\\
Parsing / Overall & 0.8823 & 0.8338\\
Knowledge / Overall & 0.9765 & 0.9235\\
\bottomrule
\end{tabular}
\end{threeparttable}
\end{table}

\subsection{Why Allocation and Sample Ranking Complement Each Other}
\label{app:within-selection}

Any size-$N$ subset $\mathcal S$ induces cluster counts
$n_k=|\mathcal S\cap\mathcal D_k|$ and shares $r_k=n_k/N$.
Its empirical sampling distribution factors as
\begin{equation}
p_{\mathcal S}(x)=r_k\,q_{\mathcal S}(x\mid k),\qquad
q_{\mathcal S}(x\mid k)=\frac{\mathbf1\{x\in\mathcal S\}}{n_k},
\quad x\in\mathcal D_k,\ n_k>0.
\label{eq:selection-factorization}
\end{equation}
Group-level allocation explicitly chooses the marginal shares $r_k$;
within-group selection determines which examples realize them. At the
finest partition, every group contains one example, and a share of
$1/N$ or zero specifies selection. Group allocation thus recovers
sample-level selection.

For intermediate partitions, composition and ranking offer complementary
control. Global top-$N$ ranking maximizes the sum of sample scores and
lets cluster shares follow from that ranking. Fixing learned quotas
instead maximizes the same scores subject to
$|\mathcal S\cap\mathcal D_k|=n_k$, solved by taking the top $n_k$
examples in each cluster. Thus, quotas control target-relevant coverage
while ranking chooses representatives within that coverage. CLIVO + ICONS
implements this factorization, combining learned composition with an
existing ranker (\Cref{tab:main-results}).

\subsection{Reusing Feedback Across Targets and Granularities}
\label{app:search-complexity}

Let $C_{\rm shared}$ denote the construction cost of embeddings,
clusters, experts, real probes, and the per-example proxy-loss cache.
The incremental cost of selecting for a new target $d$ is
\begin{equation}
\Delta C_{\rm target}(d)
=C_{\rm score}(Y,d)+C_{\rm fit}(Y,H,m_d)+C_{\rm grad}(G,T,m_d).
\label{eq:target-incremental-cost}
\end{equation}
The terms are benchmark evaluation of the $Y$ saved probes, example
screening and weight fitting on cached losses, and $T$ search updates
with $G$ adjustable shares and $m_d$ active validation examples.
Expert/probe training and cache construction are reused. Final model
training is a separate cost common to subset-selection methods.

For a new granularity at a fixed target, the incremental cost is
$\Delta C_{\rm gran}(G)=C_{\rm tie}(G)+C_{\rm grad}(G,T,m_d)$,
where $C_{\rm tie}$ constructs the parent-to-child mapping.
On the fixed $K_0=100$ expert bank, write
$D=[\Delta_1,\ldots,\Delta_{K_0}]$ and $\mathbf w=B_G\mathbf z$.
A single accumulated backward pass supplies
$\nabla_{\mathbf z}J_d=B_G^\top D^\top\nabla_\theta J_d$.
For our implementation, the search cost decomposes as
\begin{equation}
\begin{split}
C_{\rm grad}(G,T,m_d)=T\big[&C_{\rm merge}(K_0)+C_{\rm FB}(m_d)\\
&+C_{\rm directions}(K_0)+C_{\rm update}(G)\big]+C_{\rm check}.
\end{split}
\label{eq:granularity-search-cost}
\end{equation}
The dominant merge, forward/backward, and expert-direction computations
are shared across the 20/50/100-share settings; only the coefficient
mapping and update change. Finer allocation therefore adds no model
backward pass per extra share and no higher-dimensional response fit.
In comparison, a regression search that constructs $Q$ new candidate
labels incurs $Q\,C_{\rm label}+C_{\rm regfit}(Q,G)$ before proposal
and recheck costs.
\Cref{app:incremental-cost} reports the measured search workloads.

\subsection{Incremental Search Budget}
\label{app:incremental-cost}

We report compute as \emph{64-H100-equivalent hours}: H100-equivalent
GPU-hours divided by 64, using the same peak-BF16 normalization across
runs. This is a common compute-budget unit. The measured workload
includes iterative merging and model forward/backward computation;
shared construction and final model training are excluded.

\paragraph{Regression reference.}
LightGBM predictors use nested budgets of 256, 512, 1,024, 2,048, and
4,096 mixtures labeled with proxy losses. Five-fold cross-validation Spearman reaches
0.935/0.909/0.898/0.885 at 4,096 labels for
General/Parsing/Knowledge/Overall. The final doubling improves these
values by 0.024/0.013/0.055/0.036, smaller than the preceding doubling
for every target. We use these 4,096 shared labels as the high-budget reference.

\begin{table}[!htbp]
\centering\small
\begin{threeparttable}
\caption{\textbf{Incremental search compute.} Gradient costs cover all
120 updates. Regression cost covers constructing the shared 4,096
proxy-loss labels. Units: 64-H100-equivalent hours.}
\label{tab:incremental-search-cost}
\begin{tabular}{@{}lr@{}}
\toprule
Workload & Compute budget\\
\midrule
Gradient: General & 0.960\\
Gradient: Parsing & 0.993\\
Gradient: Knowledge & 1.123\\
Gradient: Overall & 0.971\\
Gradient: four-target total & 4.046\\
Regression: 4,096 shared labels & 5.509\\
\bottomrule
\end{tabular}
\end{threeparttable}
\end{table}

\Needspace{135pt}
\setlength{\intextsep}{0pt}
\begin{wraptable}{r}{0.55\textwidth}
\centering\small
\setlength{\tabcolsep}{4pt}
\begin{threeparttable}
\caption{\textbf{Search endpoints.} Single-seed 100K scores. GD:
gradient search; regression: 4,096 labeled mixtures.}
\label{tab:search-endpoints}
\begin{tabular*}{\linewidth}{@{\extracolsep{\fill}}lrrrr@{}}
\toprule
Target & GD 20 & GD 50 & GD 100 & Regression\\
\midrule
General & 67.20 & 66.83 & 67.02 & 67.09\\
Parsing & 79.81 & 80.12 & 80.26 & 80.10\\
Knowledge & 61.42 & 61.57 & 61.83 & 62.17\\
Overall & 67.99 & 67.85 & 68.02 & 68.22\\
\bottomrule
\end{tabular*}
\end{threeparttable}
\end{wraptable}
\noindent
The four gradient searches use 26.6\% less normalized compute than
regression labeling. In \Cref{tab:component-results}, $C^\star$ measures
cost to the first lowest-loss checkpoint. Upper endpoints summarize
reconstructed timing bounds; $\geq$ denotes incomplete timing. Parsing
with 20 shares retains its start, but the complete search costs
0.312 hours in the same normalized units.

\paragraph{Regression endpoint protocol.}
Regression ranks 10 million proposals by predicted log loss. Its 4,096
labels cover 115 feasible probes, four gradient endpoints, 994
perturbations of gradient solutions, and 2,983 Dirichlet samples.
Gradient endpoints contribute labels but are excluded from final
selection. \Cref{tab:search-endpoints} reports single-seed scores for
the selected candidates.

\WFclear

\end{document}